\documentclass{llncs}

\usepackage{graphicx} 
\usepackage{amssymb} 

\newcommand{\Real}{\mathop{\rm I\kern-.2emR}}
\newcommand{\V}{{\cal N}}

\begin{document}

\title{On the Neutrality of\\Flowshop Scheduling Fitness Landscapes}

\author{Marie-El\'eonore Marmion\inst{1,2} \and Clarisse~Dhaenens\inst{1,2} \and Laetitia Jourdan\inst{1} \and \\ Arnaud Liefooghe\inst{1,2} \and S\'ebastien Verel\inst{1,3}}

\authorrunning{M.-E. Marmion, C. Dhaenens, L. Jourdan, A. Liefooghe and S. Verel}

\institute{INRIA Lille-Nord Europe, France
\and Universit\'e Lille 1, LIFL -- CNRS, France
\and University of Nice Sophia Antipolis -- CNRS, France \\
\email{marie-eleonore.marmion@inria.fr, clarisse.dhaenens@lifl.fr, laetitia.jourdan@inria.fr,
arnaud.liefooghe@univ-lille1.fr, verel@i3s.unice.fr}
}

\maketitle

\begin{abstract}
Solving efficiently complex problems using metaheuristics, and in particular local search algorithms, requires incorporating knowledge about the problem to solve. In this paper, the permutation flowshop problem is studied. It is well known that in such problems, several solutions may have the same fitness value. As this neutrality property is an important issue, it should be taken into account during the design of search methods. Then, in the context of the permutation flowshop, a deep landscape analysis focused on the neutrality property is driven and propositions on the way to use this neutrality in order to guide the search efficiently are given.
\end{abstract}

\section{Motivations}

Scheduling problems form one of the most important class of combinatorial optimization problems.
They arise in situations where a set of operations (tasks) have to be performed on a set of resources (machines), optimizing a given quality criterion. 
Flowshop problems constitute a special case of scheduling problems in which an operation must pass through all the set of resources before being completed.
Such scheduling problems are often difficult to solve, because of the large search space they induce, and then represent a great challenge for combinatorial optimization.
Therefore many optimization methods have been proposed so far and experimented on a set of widely-used benchmark instances.
Regarding, the minimization of makespan in flowshop problems,
iterated local search (ILS) approaches seem to achieve very good performance.
In particular, St\"utzle's ILS \cite{stutzle:1998} stays one of the references of the literature.
It has been listed as one of the best performing metaheuristics on a review of heuristic approaches for the flowshop problem investigated in the paper~\cite{ruiz2005}.
More recently, Ruiz and St\"utzle \cite{ruiz_stutzle:2006} have proposed an iterated greedy algorithm to solve the flowshop problem, based on similar mechanisms, and they have shown that is outperforms the classical metaheuristics for this problem.

The aim of the paper is to analyze characteristics of the flowshop problems in order to understand and to explain why St\"utzle's method achieves such good performance.
A quick analysis shows that the neutrality is high in those problems and we want to explain how this neutrality influences the behavior of heuristic  methods.
It will then become possible to propose mechanisms that are able to exploit this neutrality.

The method proposed by St\"utzle consists of an Iterated Local Search (ILS) approach based on the {\it insertion} neighborhood operator.
This operator is argued to be the best one by the original author, as it produces better results than the {\it transpose} operator, for example,
while allowing a faster evaluation compared to the {\it exchange} operator.
The method starts from a solution constructed using a greedy heuristic (the NEH heuristic), initially proposed by Nawaz et al. \cite{nawaz1983}.
Next, the local search algorithm, based on a {\it first improvement} exploration of the neighborhood, is iterated until a local minimum is reached.
Then, between each local search,
a small perturbation is applied on the current solution using random applications of the {\it transpose} and {\it exchange} neighborhood operators.  
An important characteristic of this approach is the acceptance criterion of the ILS algorithm, which is based on the Metropolis condition (as in  simulated annealing).
Indeed, such a condition allows to accept a solution with a same or worse fitness value than the current one.

Hence, the contributions of this work are the following ones.
On the one hand, the specific problem of flowshop scheduling is deeply studied in terms of landscape analysis and neutrality.
On the other hand, some propositions are drawn in order to exploit neutrality in the design of a local search algorithm.
Of course, these considerations are still valid for other combinatorial optimization problems with a neutrality.

The paper is organized as follows.
Section~\ref{sec:background} is dedicated to the presentation of the flowshop scheduling problem investigated in this paper,
and of the required notions about neutrality analysis in fitness landscapes.
Section \ref{sec:analysis} presents the neutral networks analysis for the permutation flowshop problem under study,
whereas Section \ref{sec:heuristic} gives some hints on how to exploit the neutrality property in order to solve such problems efficiently by means of local search algorithms.
Finally, the last section is devoted to discussion and future works.

\section{Background}
\label{sec:background}

\begin{table}[t]
\caption{Notations used in the paper.}
\begin{center}
\begin{tabular}{r|l}
Notation & Description \\
\hline
$S$ & Set of feasible solutions in the search space \\
$s$ & A feasible solution $s \in S$ \\
$C_{max}$ & Makespan \\
$N$ & Number of jobs \\
$M$ & Number of machines \\
$\{J_1,J_2,\dots,J_N\}$ & Set of Jobs \\
$\{M_1,M_2,\dots,M_M\}$ & Set of Machines \\
$\{t_{i1},t_{i2},\dots,t_{iM}\}$ & Tasks \\
$\{p_{i1},p_{i2},\dots,p_{iM}\}$ & Processing times \\
$\{C_{i1},C_{i2},\dots,C_{iM}\}$ & Completion dates \\ 
\end{tabular}
\end{center}
\label{tab:param}
\end{table}%

\subsection{Definition of the Permutation Flowshop Scheduling Problem}
The Flowshop Scheduling Problem (FSP) is one of the most investigated scheduling problem from the literature.
The problem consists in scheduling $N$~jobs $\{J_1,J_2,\dots,J_N\}$ on $M$ machines $\{M_1,M_2,\dots,M_M\}$.
Machines are critical resources, {\itshape i.e.} two jobs cannot be assigned to the same machine at the same time.
A job $J_i$ is composed of $M$~tasks $\{t_{i1},t_{i2},\dots,t_{iM}\}$, 
where~$t_{ij}$ is the $j^{th}$ task of~$J_i$, requiring machine~$M_j$.
A processing time~$p_{ij}$ is associated with each task~$t_{ij}$.
We here focus on a permutation FSP, where the operating sequences of the jobs are identical and unidirectional for every machine.
As consequence, a feasible solution can be represented by a permutation $\pi_N$ of size $N$ (the ordered sequence of scheduled jobs),
and the size of the search space is then $|S| = N!$.

In this study, we will consider that the makespan, {\itshape i.e.}~the total completion time, is the objective function to be minimized.
Let  $C_{ij}$ be the completion date of task~$t_{ij}$, the makespan ($C_{max}$) can be computed as follows:
$$C_{max} = \max_{i \in \{1, \dots, N\}} \{C_{iM}\}$$
According to Graham et al.~\cite{graham1979}, the problem under study can be denoted by $ F / perm / C_{max} $.
The FSP can be solved in polynomial time by the Johnson's algorithm for two machines~\cite{johnson1954}. 
However, in the general case, minimizing the makespan has been proven to be NP-hard for three machines and more~\cite{lenstra1977}.
As a consequence, large-size problem instances can generally not be solved to optimality,
and then metaheuristics may appear to be good candidates to obtain well-performing solutions.

\paragraph{Benchmark Instances.}
Experiments will be driven using a set of benchmark instances originally proposed by Taillard \cite{taillard:1993} and widely used in the literature \cite{stutzle:1998,ruiz2005}.
We investigate different values of the number of jobs $N \in \{ 20,50,100,200 \}$ and of the number of machines $M \in \{ 5,10,20 \}$.
The processing time $t_{ij}$ of job $i \in N$ and machine $j \in M$ is generated randomly, according to a uniform distribution $\mathcal{U} ([0;99])$. 
For each problem size ($N \times M$), ten instances are available.
Note that, as mentioned on the Taillard's website\footnote{\url{http://mistic.heig-vd.ch/taillard/problemes.dir/ordonnancement.dir/ordonnancement.html}},
very few instances with $20$ machines have been solved to optimality.
For $5$- and $10$-machine instances, optimal solutions have been found, requiring for some of them a very long computational time.
Hence, the number of machines seems to be very determinant in the problem difficulty.
That is the reason why the results of the paper will be exposed separately for each number of machines.

\subsection{Neighborhood and Local Search}
The design of local search metaheuristics requires a proper definition of a neighborhood structure for the problem under consideration.
A \emph{neighborhood structure} is a mapping function $\mathcal{N}: S \rightarrow 2^S$ 
that assigns a set of solutions $\mathcal{N}(s) \subset S$ to any feasible solution $s \in S$.
$\mathcal{N}(s)$ is called the \emph{neighborhood} of $s$, 
and a solution $s' \in \mathcal{N}(s)$ is called a \emph{neighbor} of $s$.
A neighbor results of the application of a \emph{move operator} performing a small perturbation to solution $s$.
This neighborhood operator is a key issue for the local search efficiency.

\begin{figure}[th]
\centering
\includegraphics[width=1.5in]{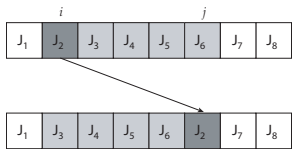}
\caption{Illustration of the \emph{insertion neighborhood} operator for the FSP.
The job located at position $i$ is inserted at position $j$, all the jobs located between $i$ and $j$ are shifted to the left.}
\label{fig:neighborhood}
\end{figure}

For the FSP, we will consider the \emph{insertion operator}.
This operator is known to be one of the best neighborhood structure for the FSP~\cite{stutzle:1998,ruiz2005}.
It can be defined as follows.
A job located at position $i$ is inserted at position~$j \neq i$.
The jobs located between positions~$i$ and~$j$ are shifted, as illustrated in Figure~\ref{fig:neighborhood}. 
The number of neighbors per solution is $(N-1)^2$, where $N$ stands for the size of the permutation (and corresponds to the number of jobs).

\subsection{Fitness Landscape}

\subsubsection{Fitness landscape with neutrality.}
In order to study the typology of problems, the fitness landscape notion has been introduced \cite{wright:1932}.
A landscape is a triplet $(S, \V, f)$ where $S$ is a set of admissible solutions ({\itshape i.e.} a search space), $\V: S \longrightarrow 2^{|S|}$, a neighborhood operator, is a function that assigns to every $s \in S$ a set of neighbors $\V(s)$, and $f: S \longrightarrow \Real$ is a fitness function that can be pictured as the \textit{height} of the corresponding solutions.
In our study, the search space is composed of permutations of size $N$ so that its size is $N!$.

\paragraph{Neutral neighbor.}
A neutral neighbor of $s$ is a neighbor solution $s'$ with the same fitness value $f(s)$. Given a solution $s \in S$, its set of neutral neighbors is defined by:
$$\V_n(s) = \{ s' \in \V(s) ~|~ f(s') = f(s) \}$$
The neutral degree of a solution is the number of its neutral neighbors.
A fitness landscape is said to be neutral if there are many solutions with a high neutral degree $|V_n(s)|$. The landscape is then composed of several sub-graphs of solutions with the same fitness value. Sometimes, another definition of neutral neighbor is used in which the fitness values are allowed to differ by a small amount. Here we stick to the strict definition given above as the fitness of flowshop (makespan) is discretized (it is an integer value).

\paragraph{Neutral network.}
A neutral network, denoted as NN, is a connected sub-graph whose vertices are solutions with the same fitness value. 
Two vertices in a NN are connected if they are neutral neighbors.
With the insertion operator, for all solutions $x$ and $y$, if $x \in \V(y)$ then $y \in \V(x)$.
So in this case, the neutral networks are the equivalent classes of the relation $R(x,y)$ iff ($x \in \V(y)$ and $f(x)=f(y)$).
We denote the neutral network of a solution $s$ by $NN(s)$.
A \textit{portal} in a NN is a solution which has at least one neighbor with a better fitness, {\itshape i.e.} a lower fitness value in a minimization context.

\paragraph{Local optimum.}
A solution $s^{*}$ is a local optimum iff  no neighbor has a better fitness value:
$\forall s \in \V(s^{*})$, $f(s^{*}) \leq f(s)$. 
When all solutions on a neutral network are local optima, the NN is a local optima neutral network.

\subsubsection{Measures of neutral fitness.}
\label{measureNeutralFit}

The average or the distribution of neutral degrees over the landscape is used to test the level of neutrality of the problem. This measure plays an important
role in the dynamics of metaheuristics \cite{NIM:99,wilke01,verel07}.
When the fitness landscape is neutral, 
the main features of the landscape can be described by its neutral networks.
Due to the number and the size of neutral networks, they are sampled by \textit{neutral walks}.
A neutral walk $W_{neut} = (s_0, s_1, \ldots, s_m)$ from $s$ to $s^{'}$ 
is a sequence of solutions belonging to $S$ where $s_0=s$ and $s_m=s^{'}$ and for all $i \in [ 0, m - 1]$ , $s_{i+1}$ is a neighbor of $s_{i}$ and $f(s_{i+1})=f(s_i)$.

A way to describe neutral networks NN is given by the \textit{autocorrelation of neutral degree} along a
neutral random walk \cite{bastolla03}. 
From neutral degrees collected along this neutral walk, 
we computed its autocorrelation function $\rho(k)$ \cite{WEI:90}, 
that is the correlation coefficient of the neutral degree between the solutions $s_{i}$ and $s_{i+k}$ for all possible $i$. 
The autocorrelation measures the correlation structure of a NN.
If the first correlation coefficient $\rho(1)$ is close to 1, the variation of neutral degree is low ;
and so, there are some areas in NN of solutions which have close neutral degrees,
which shows that NN are not random graphs.

Another interesting information to determine if a local search could find a better solution on a neutral network, is the position of portals.
The number of steps before finding a portal during a neutral random walk is a good indicator of the probability to find better solution(s) according to the computational cost to find it, {\itshape i.e.} the number of evaluations.

Moreover, to design a local search which explores the neutral networks in an efficient way, 
we need to find some information around the NN where, \textit{a priori}, there is a lack of information.
\textit{Evolvability} is defined by Altenberg \cite{kinnear:altenberg} as "the ability of random variations to sometimes produce improvement". 
The concept of evolvability could be difficult to define in combinatorial optimization. 
For example, the evolvability could be the minimum fitness which can be reached in the neighborhood.
In this work, we choose to define the evolvability of a solution as the average fitness in its neighborhood. 
It gives the expectation of fitness reachable after a random move.
The \textit{autocorrelation of evolvability} \cite{VEREL:2006:HAL-00164917:1} allows to measure the information around neutral networks.
This autocorrelation is the autocorrelation function of a evolvability measure collected during a neutral random walk.
When this correlation is large, 
the solutions which are close from each other on a neutral network have evolvabilities which are close too.
So, the evolvability could guide the search on neutral networks 
such as the fitness guides the search in the landscape where the autocorrelation of fitness values is large~\cite{WEI:90}.





\section{Neutral Networks Analysis for the Permutation Flowshop Scheduling Problem}
\label{sec:analysis}

\subsection{Experimental Design}
To analyze neutral networks, for each instance of Taillard's benchmarks, 30 different neutral walks were performed.
The neutral walks all start from a local optimum.
It has been obtained by a steepest descent algorithm initialized with a random solution.
The length of each neutral walk depends on the length of the descents which lead to local optima.
We consider 10 times the maximal length found on the 30 descents.
In the following, the results are presented according to the number of jobs~($N$) and the number of machines ($M$).
For each problem size, an average value and the corresponding standard deviation are represented.
By the term size, we mean both the number of jobs ($N$) and the number of machines ($M$).
This average value is computed from the means obtained from the 10 instances of the same size, themselves calculated from the values given by the 30 neutral walks.

\subsection{Neutral Degree}
\label{sectionNeutralDegree}

\begin{figure}[t]
\centering
\includegraphics[scale=0.40,angle=270]{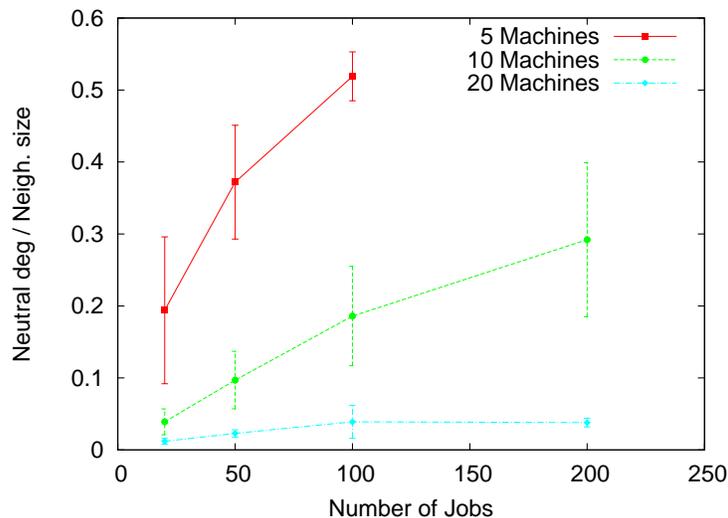}
\caption{Average of the neutral degree to the neighborhood size according to the number of jobs.}
\label{neutralDegtoNghSizePlot}
\end{figure}

\begin{figure}[t]
\centering
\includegraphics[scale=0.40,angle=270]{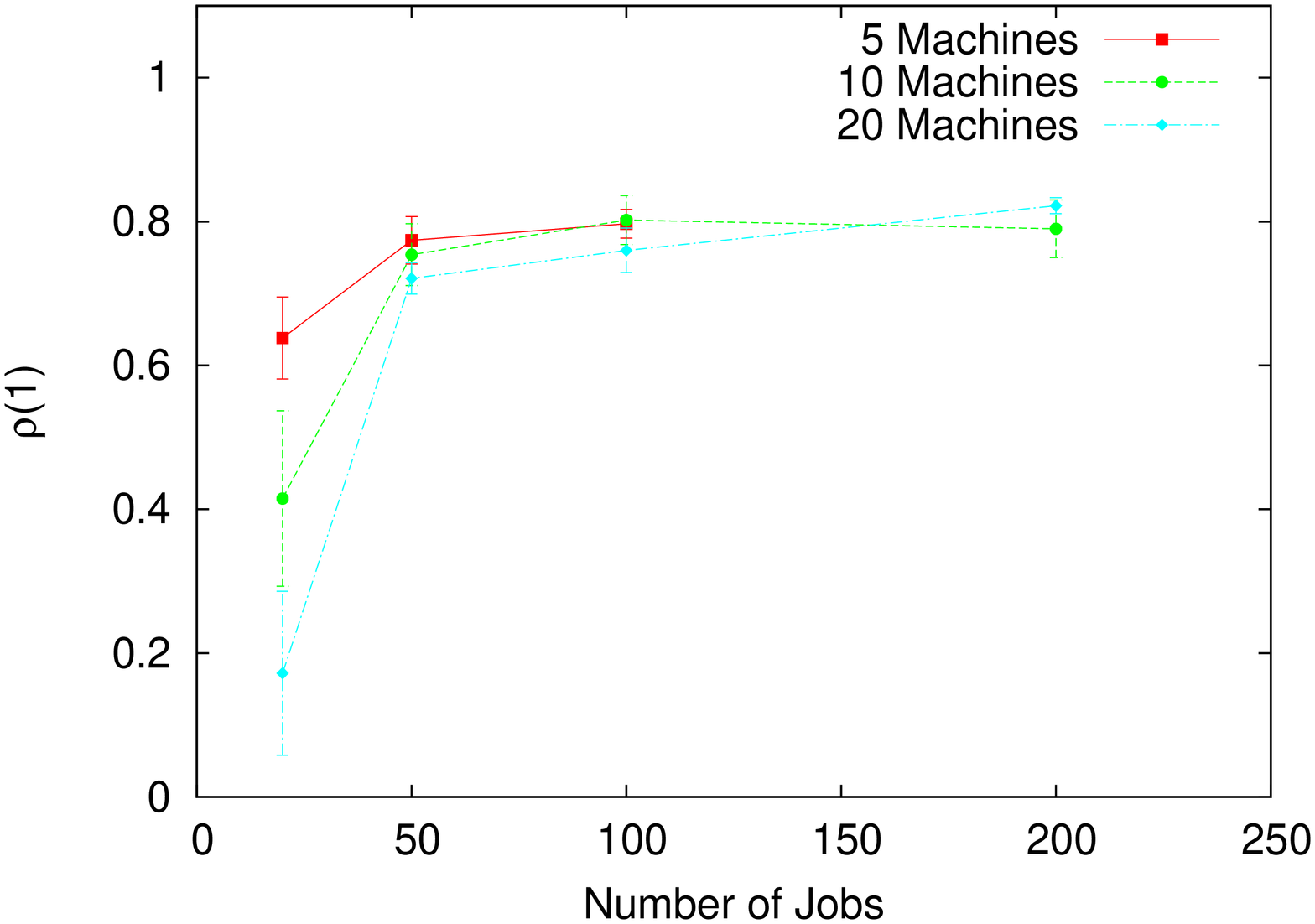}
\caption{First autocorrelation coefficient $\rho(1)$ computed between $s_i$ and $s_{i+1}$ of the neutral degree according to the number of jobs.}
\label{neutralDegreePlot}
\end{figure}

In this section, we first measure the neutral degree of the FSP.
Then, we describe the structure of the neutral networks (NN).

Figure~\ref{neutralDegtoNghSizePlot} shows the average neutral degree to the size of the neighborhood $(N-1)^2$, collected along the 30 neutral walks.
Whatever the number of machines, the neutral degree ratio increases when the number of jobs increases.
This ratio is higher for small number of machines.
For $5$-machine, and for $100$- or $200$-job and $10$-machine instances, the neutral degree is huge, higher than $20\%$.
For $100$- or $200$-job and $20$-machine instances, the ratio seems to be very low ($3.9\%$), but the number of neighbors with same fitness value is significant
(about $382$ and $1544$ neutral neighbors for $100$ and $200$ jobs, respectively).
There is no local optimum without a neighbor with the same fitness value, which means that each local optimum belongs to a local optima neutral network.
The neutral degree is high enough to describe the fitness landscape with neutral networks.

A neutral walk corresponds to a sequence of neighbor solutions on a NN of the fitness landscape, where all solutions share the same fitness value.
During those neutral walks, we compute the autocorrelation of the neutral degree (see Section \ref{measureNeutralFit}).
Figure~\ref{neutralDegreePlot} shows the first autocorrelation coefficient for $5$, $10$ and $20$ machines with respect to the number of jobs.
In order to prove that those correlations are significative, we compare them to a null model.
It consists of shuffling the same values of neutral degrees collected during the neutral walks.
Then, the autocorrelation of this model is compared to the original one.
For all sizes, the first autocorrelation coefficient of the null model is below $0.01$.
Therefore, we can conclude that the autocorrelation is a consequence of the succession of solutions encountered during the walk.

Obviously, for $50$, $100$ and $200$ jobs, the neutral degree is highly correlated (higher than $0.7$).
Moreover, the standard deviations are very low, which indicates that the average values reflect properly this property on instances of same size.
For $20$-job and $5$- or $10$-machine instances, the standard deviation gets higher.
This can possibly be explained by a higher correlation.

Nevertheless, these values allow us to conclude that the neutral degree of a solution is partially linked to the one of its neighbor solutions.
Let us remark that the correlation for $20$-job $20$-machine instances is very low, due to the small average value of the neutral degree for this size.

The first conclusions of this analysis is that
($i$) there exists a high neutrality over the fitness landscape, particularly for large-size instances
($ii$) the neutral networks, defined as the graphs of neighbor solutions with the same fitness value, are not random.
As a consequence, we should not expect to explore the neutral networks efficiently with a random walk.
Hence, heuristic methods should exploit the information available in the neighborhood of the solutions.

\subsection{Typology of Neutral Networks}

\begin{figure}[t]
\centering
\includegraphics[scale=0.70]{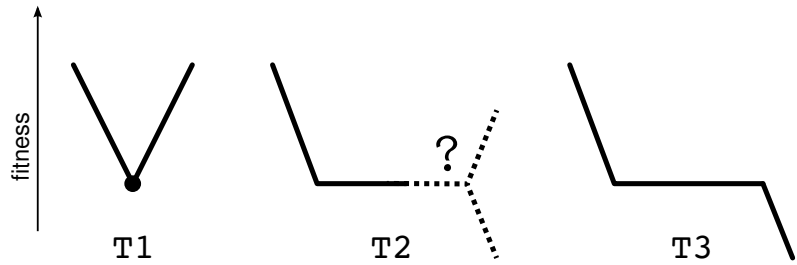}
\caption{Typology of neutral networks (minimization problem).}
\label{plateauGeometry}
\end{figure}

A metaheuristic such as ILS visits several local optima. In the previous section, we have seen that the local optima often belong to a NN.
A natural question arises when the metaheuristic reaches a NN: Is it possible to escape from this NN? 
In this section, we classify the local optima NN in three different types, and we analyze their size. 

Three types of NN typologies may exist (see Figure~\ref{plateauGeometry}):
\begin{enumerate}
\item The local optimum is the single solution on the NN (type T1), {\itshape i.e.} it has no neighbor with the same fitness value, we call it a degenerated NN.
\item The neutral walk from the local optimum did not show any neighbors with a better fitness values for all the solutions encountered along the neutral walk~(type T2).
Of course, as the whole NN has not been enumerated, we can not decide if it is possible to escape from them.
\item At least one solution having a neighbor with better fitness value than the local optimum fitness is found along the neutral walk (type T3). 
\end{enumerate}

\begin{figure}[t]
\begin{minipage}[t]{0.40\linewidth}
\centering
\includegraphics[scale=0.19,angle=270]{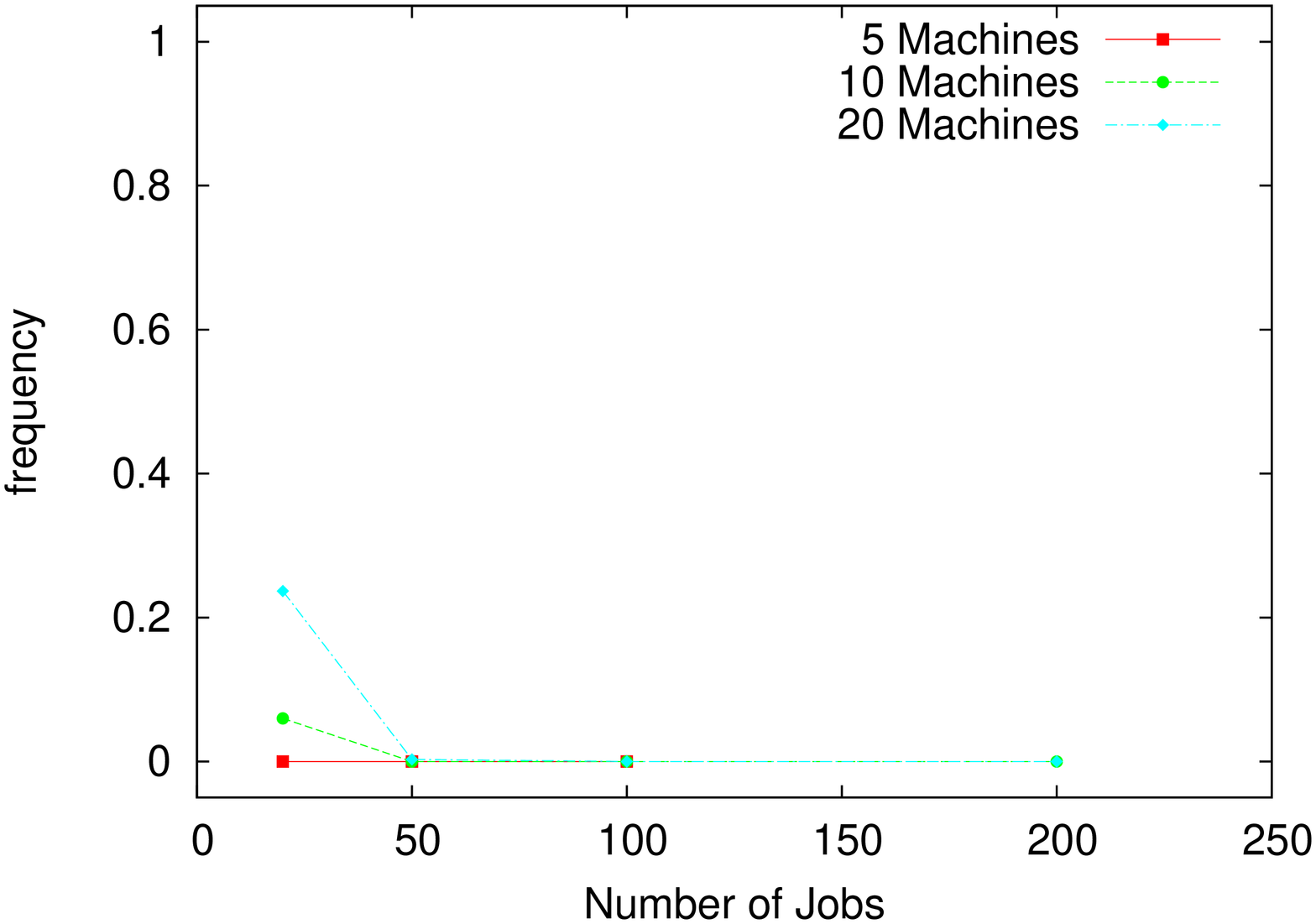}
\caption{Average frequency of the number of degenerated neutral networks with a single solution (type T1) according to the number of jobs.}
\label{percentT1}
\end{minipage} \hfill
\begin{minipage}[t]{0.40\linewidth}
\centering
\includegraphics[scale=0.19,angle=270]{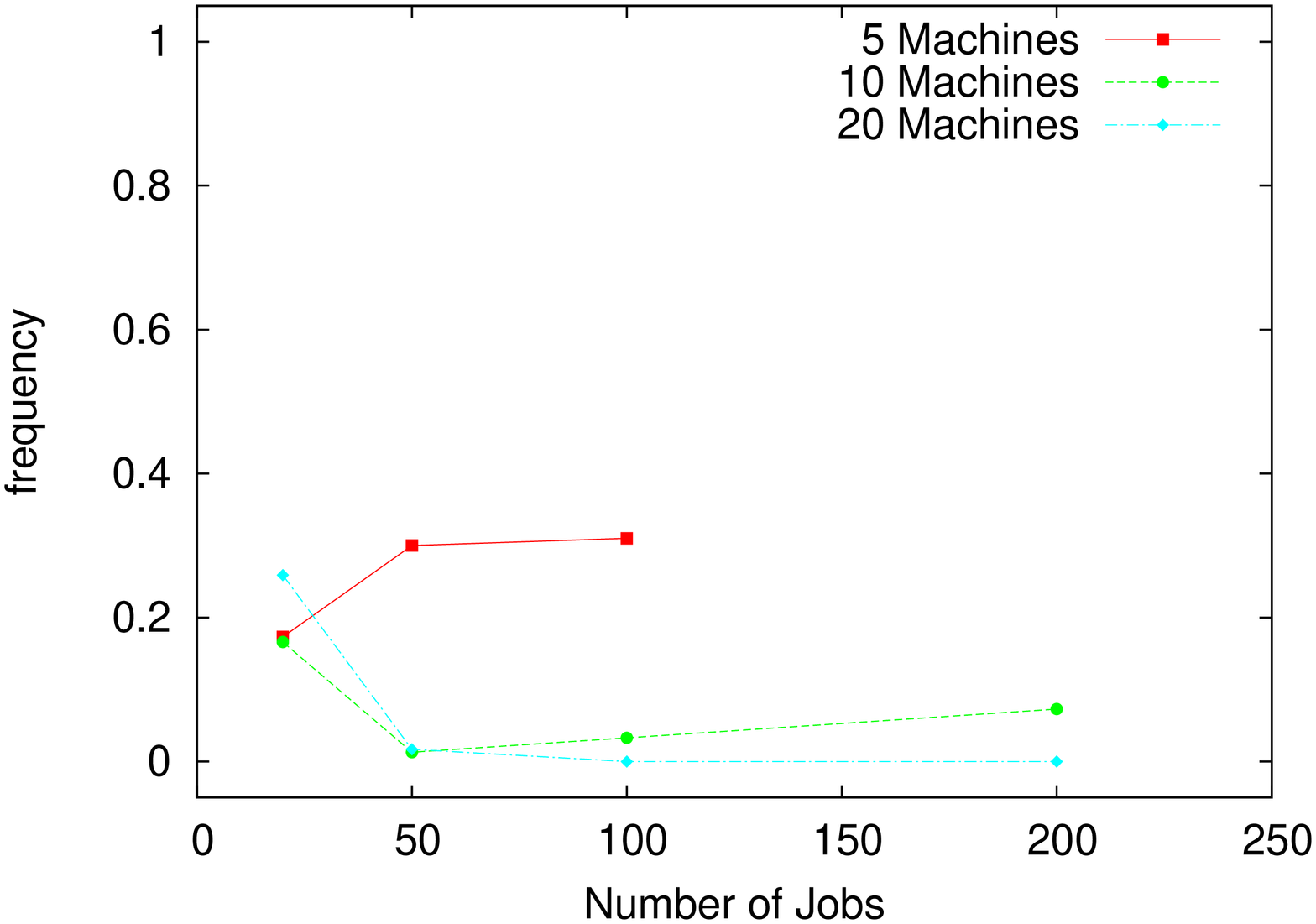}
\caption{Average frequency of the number of neutral networks where no portal was found (type T2) according to the number of jobs.}
\label{percentT23}
\end{minipage} \\
\begin{minipage}[t]{0.40\linewidth}
\centering
\includegraphics[scale=0.19,angle=270]{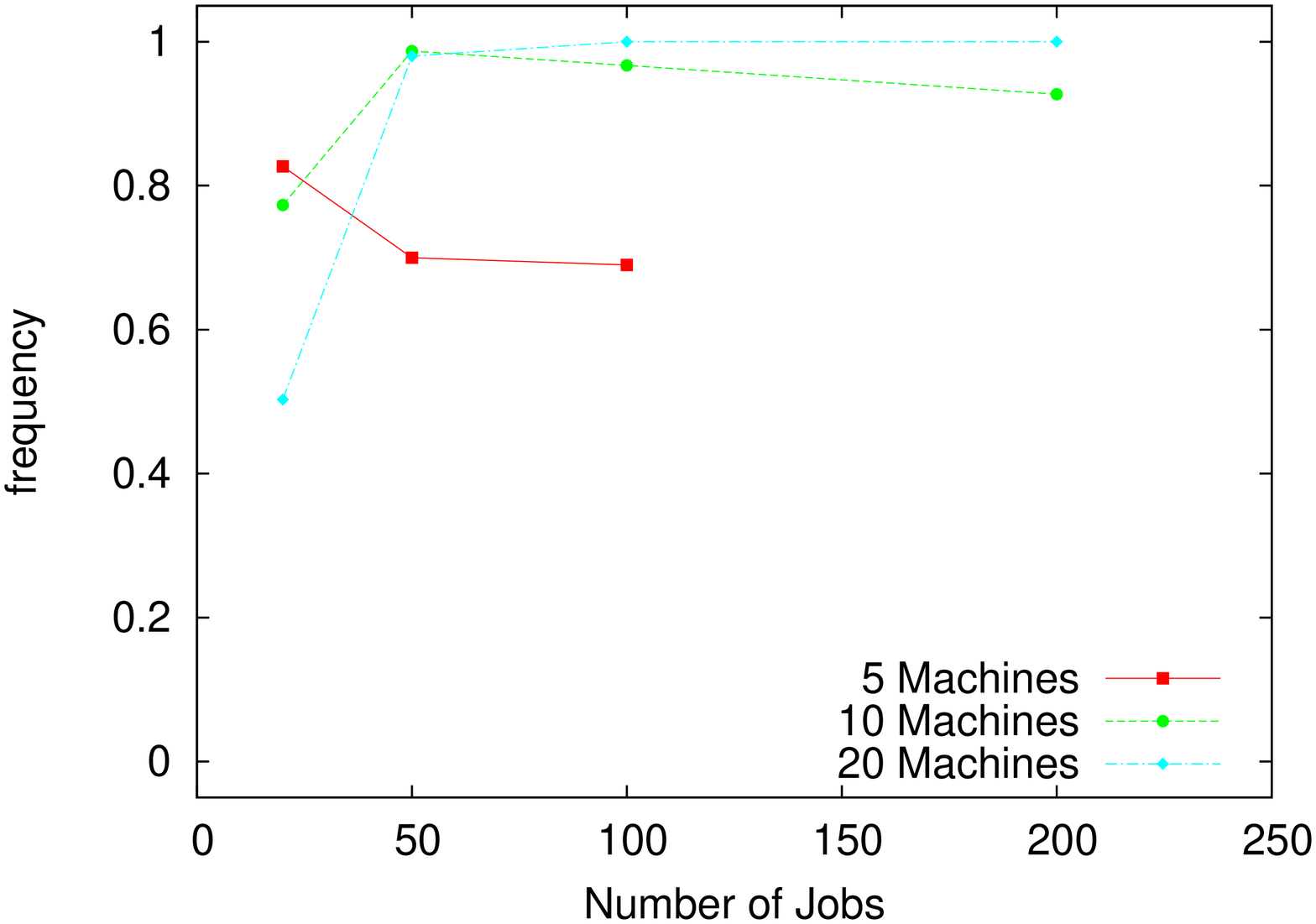}
\caption{Average frequency of the number of neutral networks where at least one portal was found (type T3) according to the number of jobs.}
\label{percentT45}
\end{minipage} \hfill
\begin{minipage}[t]{0.40\linewidth}
\centering
\includegraphics[scale=0.19,angle=270]{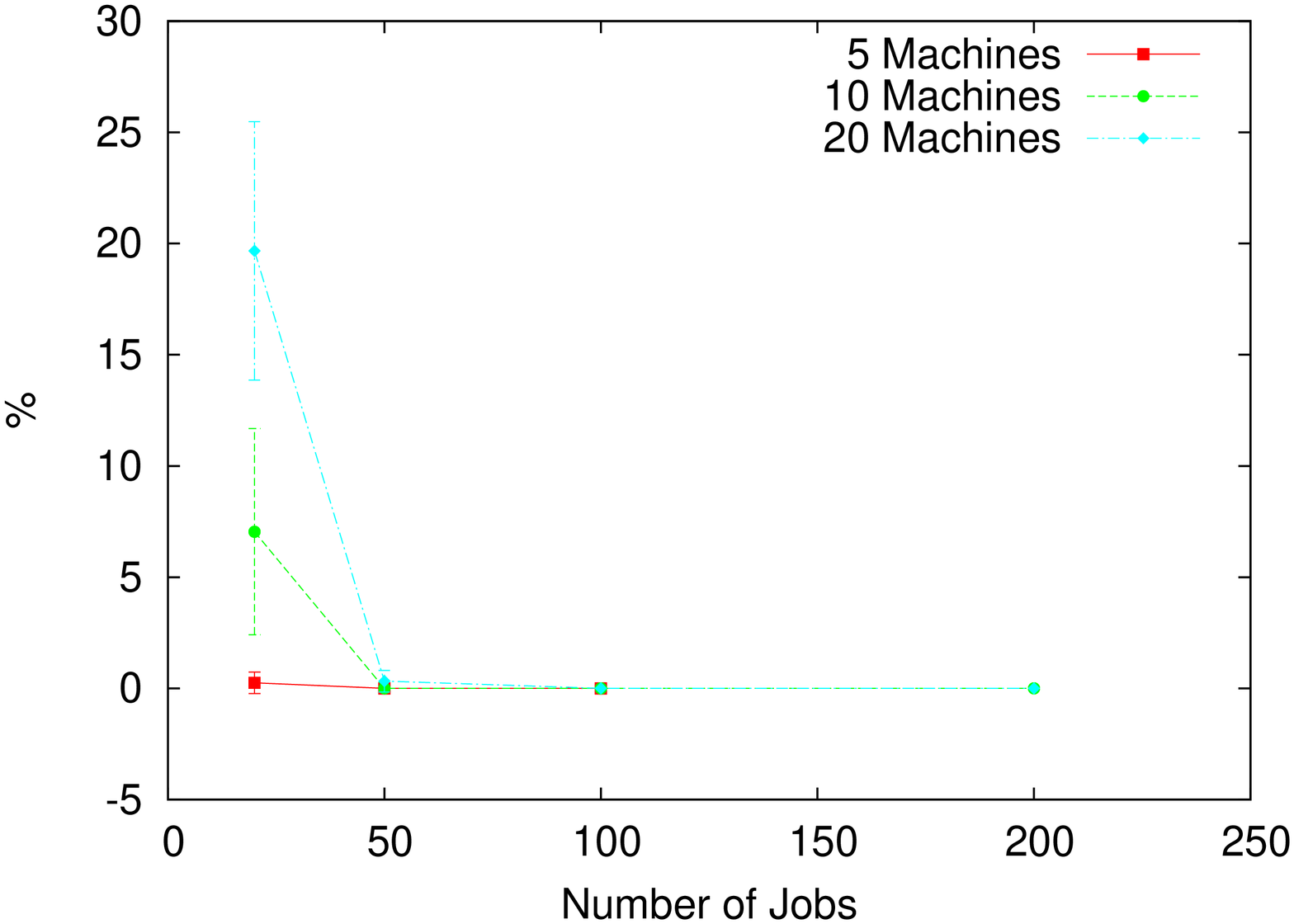}
\caption{Average percentage of solutions visited at least twice along the neutral walk according to the number of jobs.}
\label{dist0}
\end{minipage}
\end{figure}

Figures \ref{percentT1}, \ref{percentT23} and \ref{percentT45} show the proportion of NN of each type (T1, T2 or T3) counted along the neutral walks.
For $50$-, $100$- and $200$-job instances, the neutral walks show only NN of types T2 and T3. No local optimum solution is alone on the NN.
For $20$-job instances, the number of type (T1) is also small, except for 20 machines ($25 \%$ of type T1).
Hence, the neutrality is important to keep in mind while solving such instances. 
The number of NN without any escaping solutions found (T2) is significative only for $5$-machine instances (higher than $18\%$) and stays very low for $10$- an $20$-machine instances (lower than $6\%$).
The $20$-machine instances, which are known to be the hardest to solve optimally,
are the ones where the probability to escape from local optimum by neutral exploration of the NN is close to one.

When the neutral networks size is very small, the number of visited solutions is very small.
Indeed, a NN of type T2 or T3 could contain very few solutions and, the neutral walk could loop on some solutions. These situations have to be considered with attention.
Figure~\ref{dist0} shows the average percentage of solutions visited more than once during the neutral walk. 
For the $50$-, $100$- and $200$-job instances, there is no re-visited solutions during the neutral walks.
For $20$-job and $20$-machine instances, the number of re-visited solutions is approximatively $20 \%$ during neutral walks on NN of type T2 or T3. 
This result points out two remarks.
First, the NN of local optima seems to be large for most instances.
Second, the number of re-visited solutions is low, which means that the probability to escape the NN of type T2 is below the inverse of the size of the neutral walk. 

In conclusion, for most instances, a metaheuristic could escape the local optimum by exploring the NN. 
The next section will show some hints on how to guide a metaheuristic on neutral networks.


\section{Exploiting Neutrality to Solve the FSP}
\label{sec:heuristic} 

In the previous section, we proposed to use neutral exploration to escape from local optimum,
as there exists solutions having neighbor(s) with a better fitness value around neutral networks.
We called those solutions, portals.
An efficient metaheuristic has to find such portal with a minimum number of evaluations.
First, we study the number of steps to reach a portal, and then we propose an insight to get information to find them quickly.

\subsection{Reaching Portals}
\label{reachPortals}

\begin{figure}[t]
\centering
\includegraphics[scale=0.4,angle=270]{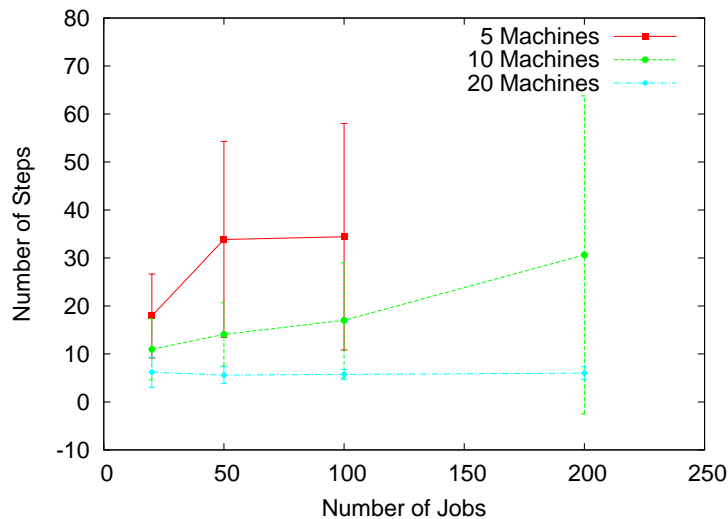}
\caption{Number of steps along the neutral random walk to reach the first portal according to number of jobs.}
\label{porteStepNb}
\end{figure}

As shown on Figure~\ref{percentT45} at least $70 \%$ of neutral random walks for FSP with 50, 100 and 200 jobs can reach a portal (more than $90 \%$ for 10 and 20 machines).
The performance of a metaheuristic which explores neutral networks highly depends on the probability to find a portal.
Indeed, it could become more time consuming to consider a neutral walk than applying a smart restart. 

Figure~\ref{porteStepNb} gives the average number of steps to reach the first portal during the 30 neutral walks. 
The larger the number of machines, the less the number of steps is required by the neutral walk to reach a portal.
For $20$-machine instances, the neutral random walks need around $7$ steps to reach a portal, which is very small compared to the length of the descents
(19, 40, 64, 101 respectively for 20, 50, 100, and 200 jobs).
For $5$-machine instances, the length of the neutral walks is around the length of the descents.
Hence, it is probably more advantageous to perform a neutral random exploration than a random restart.
Moreover, the fitness value obtained after the neutral walk is better than after the descent.
Consequently, if an \textit{a priori} study highlights that a portal is supposed to be encountered quickly,
a metaheuristic that takes into account information on the neutral walk should move on the NN, and then finally find an improving solution.

\subsection{How to Guide the Search?}
In the previous section,
the role of neutrality was demonstrated by the correlation of the neutral degree between the neutral walk neighbors and the high frequency of neutral networks.
Neutral networks lead, with very few steps, to a portal. 
The neutrality could give interesting information about the landscape in order to guide the search.
However, since the neutral network is large, the search has to be guided to find quickly a portal and not to stagnate on the NN. 
Thus, proper information has to be collected and interpreted along the neutral walk to help the metaheuristic to take good decision: 
Is it more interesting to continue the neutral walk until a portal is reached or to restart? 
As suggested in Section \ref{measureNeutralFit}, we compute the evolvability of a solution as the average fitness values of its neighbors for all visited solutions.
We analyze the evolvability of solutions on neutral networks and we give some results about the correlation of evolvability and portals on a neutral network.
This allows us to propose new ideas for the design of a metaheuristic.

\begin{figure}[t]
\centering
\includegraphics[scale=0.40,angle=270]{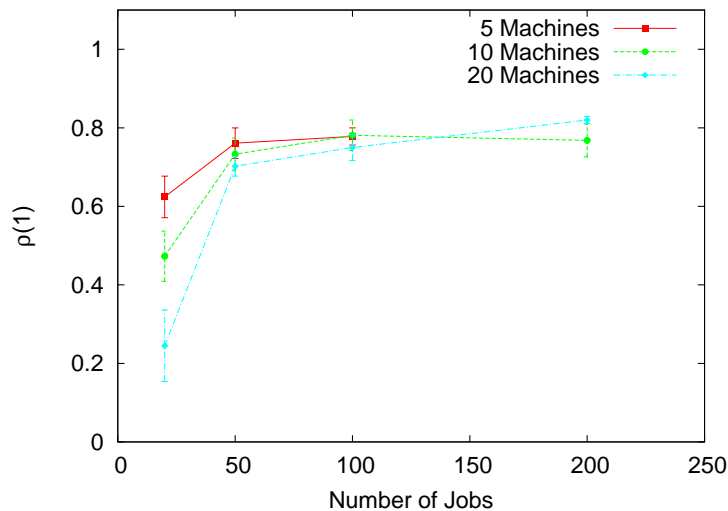}
\caption{First autocorrelation coefficient $\rho(1)$ of the average fitness values of neighbors solutions between $s_i$ and $s_{i+1}$ according to the number of jobs.}
\label{correlMeanFitNghPlot}
\end{figure}

During those neutral walks, we compute the evolvability of each solution along the neutral walk, and then its autocorrelation (see section \ref{measureNeutralFit}).
Figure~\ref{correlMeanFitNghPlot} shows the first autocorrelation coefficient $\rho (1)$ for $5$, $10$ and $20$ machines with respect to the number of jobs.
In order to show that those correlations are significative, as in section \ref{sectionNeutralDegree}, we compare them to a null model.
For all sizes, the first autocorrelation coefficient of the null model is below $0.01$.
Therefore, we can conclude that the autocorrelation is a consequence of the succession of solutions encountered during the walk.
The average fitness values of the neighbors are not distributed randomly: they can then be exploited by a metaheuristic.

\begin{figure}[t]
\centering
\includegraphics[scale=0.40,angle=270]{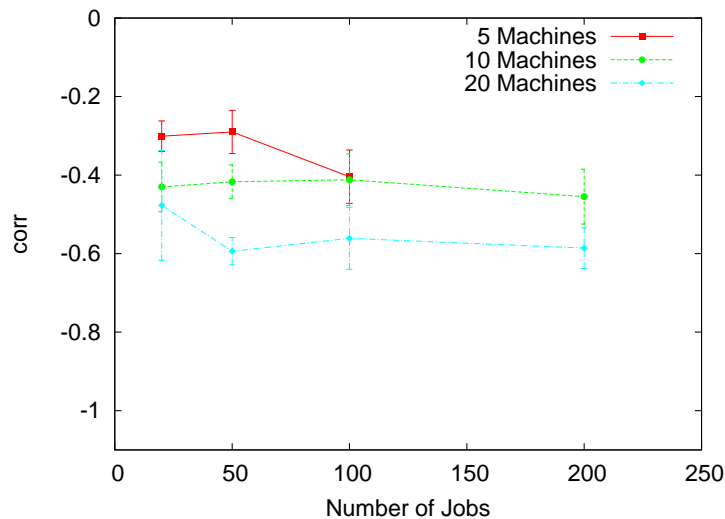}
\caption{Correlation between the average fitness values of the neighbors and the number of steps required to reach the closer portal  according to the number of jobs.}
\label{correlMeanFitNbImpNgh}
\end{figure}

The neutral networks present evolvability and portals.
So, we can wonder if the evolvability would be able to guide a metaheuristic quickly to a portal. 
To test this hypothesis, along the neutral walks, we compute the correlation between the average fitness values in the neighborhood and the number of steps required to reach the closer portal of the walk.
This is presented in Figure~\ref{correlMeanFitNbImpNgh}. 
The larger the number of machines, the higher (in absolute value) the negative correlation.
For $10$- or $20$-machine instances, this correlation belongs to $[-0.6;-0.4]$, so that it is significant for a metaheuristic to use such an information.
The lower the average fitness values in the neighborhood, the closer a portal is.
Consequently, we propose to design a metaheuristic that takes into account the neutrality by allowing the exploration of solutions along the neutral walk. 
Starting from a local optimum, it would choose the next neutral solution with the lower average fitness values of its neighbors.
This would increase the probability to find a portal quickly, and then to continue the search process.

\section{Discussion}

In this work, we studied the neutrality of the FSP on a set of benchmark instances originally proposed by Taillard. 
Most of the instances have a high neutral degree: for a solution, the number of its neighbors with the same fitness value is significant in comparison to the neighborhood size.
Starting from local optima, neutral walks have been performed. 
Each walk moves from a solution to another with the same fitness value and defines a neutral network that is shown to be structured. Indeed, the graph of neighbor solutions is not random and so a solution shares information with its neighbors.
We show that a neutral walk leads easily to portals, solutions of the neutral network having a neighbor with a better fitness value.
Furthermore, the evolvability, defined in this study as the average fitness values of the neighbors, is highly autocorrelated. It proves that this information is not random between the neighbor solutions and so it could be helpful to take it into account. 
Besides, improving the evolvability during the neutral walk often leads to a portal. 
This work completes the knowledge of FSP fitness landscape, and in particular, about its neutrality.
Here, the neutrality has been shown for the FSP Taillard instances where the durations of jobs are integer values from $[0;99]$. 
This is a specific choice which could have an impact on the difficulty of instances.
Future works will consider other instance generators, and study the neutrality according to the instance parameters.

This work also helps to understand some experimental results on the efficiency of metaheuristics.
In a study of iterated local search to solve the FSP \cite{stutzle:1998}, St\"utzle designs several efficient ILS, called ILS-S-PFSP and compares them to local search algorithms.
He writes: "Experimentally, we found that rather small modifications [of the solution] are sufficient to yield very good performance". 
In section \ref{reachPortals}, we show that improving solutions can be reached very quickly applying \textit{insertion} operator on a neutral network. 
So, St\"utzle's remark can be explained by the neutrality and the high probability on the neutral networks to move on a solution with an improving neighbor.
Moreover, this works supports the experimentations on ILS design for $20$-machine instances.
The study of neutral walks highlights features that explain the efficient design of the ILS-S-PFSP.
Indeed, remember that the ILS-S-PFSP, initialized with a random solution, applies a local search based on \textit{insertion}-neighborhood mapping to get a local optimum, and then applies iteratively the steps ($i$) perturbation, ($ii$) local search, and ($iii$) acceptance criterion, until a termination condition is met. 
All acceptance criteria tested in ILS-S-PFSP are based on the Metropolis condition: they always accept a solution with equal fitness value.
So the neutral moves are always accepted.
Besides, St\"utzle work shows that the perturbation based on the application of several \textit{swap} operators (also called \textit{transpose} operators) is efficient.
And, the \textit{swap} neighborhood is included in the \textit{insertion} neighborhood as the job $i$ can be inserted at the positions $(i-1)$ or         $(i+1)$.
So, applying the \textit{swap} operator several times could correspond to a walk on a neutral network defined by \textit{insertion}-neighborhood relation.
Thus, steps ($i$) and ($iii$) allow the ILS-S-PFSP to move on the neutral network that could be frequent for those FSP instances.
Moreover, we show that the distance is small between a local optimum and a portal. 
So, such an ILS-S-PFSP is able to quickly improve the current best solution, which could explain its performances.

Furthermore, our work proposes to consider the neutrality to guide a metaheuristic on the search space. 
The FSP instances shows neutrality, it is easy to encounter portals along a neutral walk and the evolvability leads quickly to them.
With such information, a metaheuristic is proposed: first a local search is performed from a random solution, and then iteratively ($i$) the evolvability on the neutral network is optimized until a portal is found and ($ii$) the local search is applied to move to an other local optimum. The metaheuristic finishes when the termination criterion is met.
Similar ideas have been ever tested on other problems with neutrality such as Max-SAT and NK-landscapes with neutrality~\cite{VEREL:2004:HAL-00160035:1}.
A first attempt for developing such a strategy leads to the proposition of NILS \cite{Marmion:EvoCop2011} that has been successfully tested on flowshop problems.

%
%
\bibliographystyle{splncs}


\end{document}